\documentclass{article}

 \usepackage[preprint]{neurips_2026}

\usepackage[utf8]{inputenc} 
\usepackage[T1]{fontenc}    
\usepackage{hyperref}       
\usepackage{url}            
\usepackage{booktabs}       
\usepackage{amsfonts}       
\usepackage{nicefrac}       
\usepackage{microtype}      
\usepackage{xcolor}         

\usepackage{multirow}
\usepackage{svg}
\usepackage{amsmath}
\usepackage{booktabs}
\usepackage{subcaption}
\usepackage{float}

\title{Interpretable and Steerable Molecular Graph Generation using Concept Bottlenecks}

\author{
Sarabeshwar Balaji\textsuperscript{1}
\And
Ritesh Kumar\textsuperscript{2,3}
\And
Lav R.~Varshney\textsuperscript{4}
}

\begin{document}

\maketitle
\begingroup
\renewcommand{\thefootnote}{}
\footnotetext{
\textsuperscript{1}Indian Institute of Science Education and Research Bhopal, India;
\textsuperscript{2}CSIR-Central Scientific Instruments Organisation, Chandigarh, India;
\textsuperscript{3}Academy of Scientific and Innovative Research (AcSIR), Ghaziabad, India;
\textsuperscript{4}Stony Brook University AI Innovation Institute, Stony Book, NY, USA. 
Correspondence to: Sarabeshwar Balaji <sarabeshwar23@iiserb.ac.in>
}

\begin{abstract}
Deep graph generators (DGGs) largely lack interpretability, which is a significant limitation for one of their most crucial applications, \emph{drug discovery}, where interpretable and transparent generation could provide greater control over the desirability of generated molecules, thereby reducing trial-and-error in candidate selection. Common approaches to interpretability, such as disentanglement, cannot guarantee that the learned representations are human-interpretable. To address this, we propose incorporating concept bottlenecks into DGGs, allowing users to \emph{interpret}, \emph{debug}, and \emph{steer} the generation process. We show that this consistently improves distributional fidelity (lower FCD), with only modest, non-systematic effects on validity and uniqueness. We further explore \emph{Concept Controller}, an optimization-based steering approach that has proven to be effective for steerable image generation. We evaluate our approach on a one-shot molecular graph generation framework, though the underlying ideas can be transferred to sequential graph generation as well. 
\end{abstract}

\section{Introduction}
Graphs are used to represent relational and structural information across diverse application domains. Consequently, graph generation has attracted significant interest in several fields \cite{guo2022systematic}. However, the generative process involved in these methods has low interpretability \cite{tann2020shadowcast, zhu2022survey}. Unlike deep generative models in the image and text domains, which have increasingly moved towards more human-interpretable paradigms, interpretability remains relatively underexplored in graph generation \cite{zhu2022survey, faez2021deep, guo2022systematic}. This gap is particularly consequential in \emph{drug discovery}, where greater transparency in the generation process affords greater control over sample desirability, reducing costly trial-and-error in candidate selection~\cite{faez2021deep}.

A common approach to encourage interpretability in generative models is to learn representation that can be decomposed into independent factors (disentanglement)~\cite{tran2017disentangled, higgins2017beta}. But these representations may not be human-interpretable~\cite{locatello2019challenging}. To address this challenge, we propose a new line of research in graph generation that uses concept bottlenecks to develop inherently interpretable generative models \cite{ismail2024concept, kulkarni2025interpretable, martinez2026probabilistic}. Such methods involve training a concept bottleneck layer which embeds the latent representation of the input as a set of human understandable concepts. These interpretable embeddings are then passed to the decoder. This allows users to \emph{interpret}, \emph{debug} and \emph{steer} the output of the model. Ismail et al.~\cite{ismail2024concept} show that such frameworks outperform conditioning models for image generation, especially when the number of concepts is large. We also explore an alternative framework (\textit{Concept Controller}~\cite{kulkarni2025interpretable}) that is trained post-hoc to predict concepts from the latent representation and uses an optimization-based intervention to steer the output.  

The main idea of this work is to adapt concept bottleneck generative models and the Concept Controller to the domain of molecular graph generation. We show that this improves distributional similarity to the reference set (lower FCD~\cite{preuer2018frechet}), at low cost to validity and uniqueness. Refer to Figure~\ref{fig:main_figure}.

\begin{figure}[h]
    \centering
    \includegraphics[width=\textwidth]{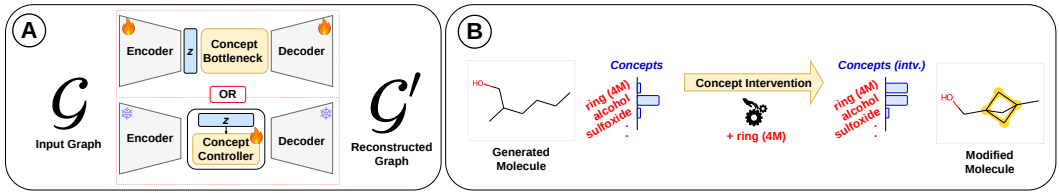}
    \caption{\textbf{(A) Training.} The concept bottleneck layer is inserted after the latent representation, and the entire generative model is trained end-to-end. Alternatively, Concept Controller (CC) can be trained post-hoc with the pretrained encoder and decoder kept frozen. \textbf{(B) Concept Intervention.} A concept of the generated molecule is intervened on to include a four-membered ring.
}
    \label{fig:main_figure}
\end{figure}

\section{Methodology}

\subsection{Steerable and Interpretable Generation}
\label{sec:cbgm}

\paragraph{Concept Bottleneck Generative Model} A concept bottleneck is an intermediate representation within a model that is constrained to align with a set of human-understandable concepts $\mathbf{c}$. Concept bottleneck models~\cite{koh2020concept} were originally proposed for classification tasks, so that the model prediction is mediated entirely through human-interpretable concepts. To the best of our knowledge, Ismail et al.~\cite{ismail2024concept} were the first to extend this idea to generation tasks, introducing concept bottleneck generative models (CBGM) to enable interpretability and steerability of generated outputs. Although their approach was demonstrated on image generation, the underlying formulation is agnostic to data modality, which motivates its use for molecular graph generation. A generative model can be viewed as an encoder-decoder pair, where an encoder maps an input $\mathbf{x}$ to a latent representation $\mathbf{z}$, i.e., $\mathbf{x} \rightarrow \mathbf{z}$, and a decoder that maps the latent representation $\mathbf{z}$ back to the data space, i.e. $ \mathbf{z} \rightarrow \mathbf{x}$. In a CBGM, a concept bottleneck layer is inserted at an intermediate point in the decoder, so that the latent vector $ \mathbf{z}$ is first mapped onto a pre-defined set of concepts $\mathbf{c}$, before being mapped to the output. This modifies the generative pathway to $\mathbf{z} \rightarrow \mathbf{c} \rightarrow \mathbf{x}$. Because the concept vector $\mathbf{c}$ is human-interpretable, CBGM supports test-time intervention: the value of a target concept $\mathbf{c_i}$ within $\mathbf{c}$ can be manually replaced by a desired value $\mathbf{c_i^*}$, resulting in an intervened concept vector $\mathbf{c^*}$. Passing $\mathbf{c^*}$ through the decoder ($\mathbf{c^*} \rightarrow \mathbf{x^*}$) produces a new modified output $\mathbf{x^*}$ in which the desired concept has been modified. In CBGM, the concept-alignment loss is optimized jointly with the backbone generative model specific loss. 

\paragraph{Concept Controller} Kulkarni et al.~\cite{kulkarni2025interpretable} propose the Concept Controller (CC), a module $\Omega$ that predicts concepts directly from the latent representation of a pre-trained generative model, i.e., $\mathbf{z} \rightarrow \mathbf{c}$. Unlike CBGM, CC is trained \emph{post-hoc}. The generative backbone is pre-trained end-to-end ($\mathbf{x} \rightarrow \mathbf{z} \rightarrow \mathbf{x}$), after which its encoder and decoder weights are kept frozen, and only $\Omega$ is trained to predict the concepts from the latent representation ($\mathbf{z} \rightarrow \mathbf{c}$). Steering is performed using a gradient-based intervention inspired by adversarial attacks. Given a latent $ \mathbf{z}$ with predicted concepts $\mathbf{c} = \Omega(\mathbf{z})$, a perturbation $\delta$ is found that steers the predicted concepts toward a target concept vector $\mathbf{c^*}$:
\begin{equation}
\mathbf{z^*} = \mathbf{z} + \arg\max_{\delta \in \Delta} \left[ -\mathcal{L}_c\big(\Omega(\mathbf{z} + \delta),\, \mathbf{c^*}\big) \right],
\end{equation}
where $\Delta = \{\delta : \|\delta\|_\infty \leq \epsilon\}$ is an $\ell_\infty$-norm ball of radius $\epsilon = 0.5$, and $\mathcal{L}_c$ is a concept alignment loss. The resulting output $\mathbf{x^*}$, obtained by passing $\mathbf{z^*}$ through the decoder, closely resembles the original generation $\mathbf{x}$ but reflects the desired target concepts $\mathbf{c^*}$.

\subsection{Graph Variational Autoencoder}
A molecule can be represented using a graph $G$, which is defined using its adjacency matrix $A$, node attribute matrix $F$ and edge attribute tensor $E$. The edge attributes one-hot encode the bond types between atoms, while $F$ one-hot encodes the atom types. We use the graph generative framework proposed by Simonovsky et al.~\cite{simonovsky2018graphvae} to learn a latent space from which graphs can be sampled. We refer to this framework as GVAE. GVAE consists of an encoder $q_\phi(\mathbf{z} \mid G)$, which outputs the parameters of the variational posterior distribution and a decoder $p_\theta(G \mid \mathbf{z})$, where $\theta$ and $\phi$ are learned parameters. Additionally, a Gaussian prior $p(\mathbf{z}) = \mathcal{N}(0, I)$ is imposed on the latent space as a regularization. This framework is trained by minimizing the upper bound on the negative log-likelihood~\cite{kingma2013auto}: 
\begin{equation}
\mathcal{L}(\phi,\theta;G)
=
\mathbb{E}_{q_\phi(\mathbf{z}\mid G)}
\left[-\log p_\theta(G\mid \mathbf{z})\right]
+
\mathrm{KL}\left[q_\phi(\mathbf{z}\mid G)\,\|\,p(\mathbf{z})\right].
\label{eq:vae_loss}
\end{equation}

\paragraph{Encoder \& Probabilistic Decoder}
The encoder is implemented using two-edge conditioned graph convolution layers~\cite{simonovsky2017dynamic} followed by graph level readout layer following the formulation in Equation 7 of Li et al.~\cite{li2015gated}. The decoder is implemented by a parallel triplet of fully connected layers to output fully connected probabilistic graph $\tilde{G} = (\tilde{A},\tilde{E},\tilde{F})$ on $k$ nodes, where $k$ is the maximum number of nodes. At test time, a discrete graph estimate is obtained from the predicted adjacency matrix $\tilde{A} \in[0,1]^{k \times k}$, the node feature matrix $\tilde{F} \in\mathbb{R}^{k \times d_n}$, and the edge feature tensor $\tilde{E} \in \mathbb{R}^{k \times k \times d_e}$ by applying an $\arg\max$ operation over the respective feature dimensions, with the adjacency matrix being discretized using a threshold of $0.5$. The prior assumption that molecular graphs are connected is furthermore incorporated at inference time. A maximum spanning tree is constructed over the set of probable nodes $\{a : \tilde{A}_{a,a} \geq 0.5\}$. Consequently, the edges of this tree are included in the discrete estimate even if their corresponding predicted probabilities $\tilde{A}_{a,b}$ fall below the threshold of $0.5$. Hydrogen atoms are not explicitly generated and are instead added implicitly as padding during the chemical validity assessment.

\paragraph{Reconstruction Loss} We adopt the factorized reconstruction loss of Simonovsky et al.~\cite{simonovsky2018graphvae}, splitting it into adjacency, node-attribute, and edge-attribute terms, but replace their graph matching step with \emph{RDKit}~\cite{landrum2016rdkit} canonical node ordering for efficiency. Full details are given in Appendix~\ref{appendix:recon}.

\section{Experiments}

\subsection{Experimental Setup}

\paragraph{Dataset \& Concept Set} We use the QM9 dataset~\cite{ramakrishnan2014quantum}, which contains about 134K organic molecules of up to 9 heavy (non-hydrogen) atoms. We set accordingly $k=9$, $d_n=4$, and $d_e=3$. We set aside 20K samples for validation and 13K samples for testing. We use structural concepts derived from the MACCS keys, as implemented in RDKit, as our vocabulary of concepts~\cite{durant2002reoptimization, landrum2016rdkit}. Each MACCS key is a deterministic bit associated with the presence or absence of a specific substructure. From the full set of keys, we select subsets of bits to define our concept sets at two sizes: 10 bits for a small, and 35 bits for a larger concept set. For each size, we further consider two variants: (i) subset of balanced bits, and (ii) subset of balanced and non-correlated bits. This yields four concept set configurations: \textit{B-10}, \textit{N-10}, \textit{B-35}, and \textit{N-35}, where \textit{B} and \textit{N} denote balanced and non-correlated, respectively. The full list of bits used in each configuration is provided in Appendix~\ref{appendix:concept_bit_list}.

\paragraph{Concept Alignment} Since all concepts are binary, we use a cross-entropy loss as the concept-alignment loss throughout. For CBGM, this loss is added directly to the generative model loss in Equation~\ref{eq:vae_loss}, following the joint end-to-end training scheme described in Section~\ref{sec:cbgm}. For the CC the same loss function is used to train $\Omega$ post-hoc.

\subsection{Generation Quality \& Interpretability}

\paragraph{Generation Quality} We evaluate the quality of generation for each paradigm by sampling $10^4$ latent vectors $\mathbf{z} \sim p(\mathbf{z})$ and generating the corresponding molecules. The metrics we use to evaluate the quality of molecule generation are validity (fraction of chemically valid molecules generated) and uniqueness (fraction of unique molecules present in valid molecules). Additionally, we utilized Fr\'echet ChemNet Distance (FCD)~\cite{preuer2018frechet, polykovskiy2020molecular} to evaluate the diversity of molecules generated. FCD is analogous to the Fr\'echet Inception Distance (FID)~\cite{heusel2017gans}, which is commonly used to quantify the distributional similarity between generated and reference images. Results are shown in Table~\ref{tab:generation_metrics_consise}. See Appendix~\ref{appendix:random_samples} for qualitative examples of generated molecules. All  GVAE with concept bottleneck (CB-GVAE) variants reduce FCD relative to GVAE, with B-35 achieving the greatest improvement. Validity and uniqueness remain broadly comparable to baseline across variants, with only modest, configuration-dependent variation. This aligns with the observation of Kulkarni et al.~\cite{kulkarni2025interpretable} and Ismail et al.~\cite{ismail2024concept} for image generation. 

\begin{table}[htbp]
\centering
\begin{minipage}[t]{0.47\textwidth}
\centering
\footnotesize
\caption{Molecular generation performance.}
\label{tab:generation_metrics_consise}
\begin{tabular}{cccc}
\toprule
Model & Val.($\uparrow$) & Uniq.($\uparrow$) & FCD($\downarrow$) \\
\midrule
GVAE            & $73.79$ & $86.53$ & $6.60$ \\
CB-GVAE (B-10)  & $69.90$ & $82.96$ & $5.28$ \\
CB-GVAE (N-10)  & $74.05$ & $83.55$ & $4.99$ \\
CB-GVAE (B-35)  & $70.32$ & $91.61$ & $4.49$ \\
CB-GVAE (N-35)  & $68.51$ & $88.81$ & $6.14$ \\
\bottomrule
\end{tabular}
\end{minipage}
\begin{minipage}[t]{0.47\textwidth}
\centering
\scriptsize
\caption{Concept accuracy.}
\label{tab:concept_accuracy}
\begin{tabular}{ccc}
\toprule
Concept Set & Model & Accuracy($\uparrow$)\\
\midrule
\multirow{2}{*}{B-10}
& CB-GVAE & $63.48$ \\
& CC-GVAE & $64.45$ \\
\midrule
\multirow{2}{*}{N-10}
& CB-GVAE & $65.55$ \\
& CC-GVAE & $67.64$ \\
\midrule
\multirow{2}{*}{B-35}
& CB-GVAE & $62.29$ \\
& CC-GVAE & $65.50$ \\
\midrule
\multirow{2}{*}{N-35}
& CB-GVAE & $71.07$ \\
& CC-GVAE & $76.66$ \\
\bottomrule
\end{tabular}
\end{minipage}
\end{table}

\paragraph{Concept Accuracy}Concept accuracy measures the average agreement between predicted concepts of the model and the ground-truth concepts present in the generated output. In Table~\ref{tab:concept_accuracy}, we report the concept accuracies of CB-GVAE and GVAE with CC (CC-GVAE). We calculate the concept accuracy over the subset of valid molecules among the $10^4$ samples generated earlier to evaluate the quality of generation. CC-GVAE outperforms CB-GVAE across all the concepts sets. This aligns with the observation by Kulkarni et al.~\cite{kulkarni2025interpretable} for image generation. We note that using uncorrelated concepts instead of balanced concepts leads to significant increase in accuracy in large concept sets than in smaller ones.

\subsection{Steerability}
\paragraph{Single-concept intervention} To evaluate the steerability for \emph{activation} of concepts, we find $10^3$ latent vectors for each concept such that the molecules generated from these vectors have the concepts absent. Similarly, to evaluate the steerability for \emph{activation} of concepts, we find $10^3$ latent vectors for each concept such that the generated molecules have the concepts present. For each steering paradigm, we intervene on the specific concept probability to 'turn-on' or 'turn-off' the concept accordingly and generate the new outputs. For each concept, we measure the fraction of inputs for which the interventions is accurately reflected in the newly generated molecule (target accuracy). We consider the intervention to be failed if the newly generated molecule is chemically invalid. Inspired by Mart{\'\i}nez-Garc{\'\i}a et al.~\cite{martinez2026probabilistic}, we also assess whether the non-target concepts remain unchanged (non-target accuracy). 

\paragraph{Minimum Hamming distance intervention} Intervening on a single concept can yield a concept configuration that is out-of-distribution. We therefore ground each intervention in the data, steering instead towards the nearest training-set configuration, in Hamming distance, that exhibits the target concept. This changes the fewest concepts needed to stay in distribution, separating failures of the concept bottleneck from failures caused by unrealizable concept configurations.

In Table~\ref{tab:concept_steerability}, we report the average steerability for several concepts sets for different modules. For a baseline, we utilize conditional version of GVAE (C-GVAE) as proposed by Simonovsky et al.~\cite{simonovsky2018graphvae} (see Appendix~\ref{appendix:c-gvae}). We show some qualitative results in Figure~\ref{fig:activation_deactivation}.

For single-concept interventions, CB-GVAE outperforms CC-GVAE on small concept sets in target accuracies, while being mostly at par with C-GVAE in the large concept set regime. Unfortunately, CC-GVAE and CB-GVAE underperform across all accuracies in non-correlated large concept set. Non-target accuracies for CB-GVAE and CC-GVAE significantly improve when using non-correlated concepts instead of balanced concepts for a given concept set size. We note that choosing the value of $\epsilon$ for CC-GVAE has a tradeoff effect between the target and non-target accuracies (see Appendix~\ref{appendix:epsilon}).

For minimum Hamming distance interventions, CB-GVAE outperforms C-GVAE across all concept sets in target accuracy for both activation and deactivation. Notably, for large concept sets, CB-GVAE achieves up to a twofold increase in target accuracy for both types of intervention. In contrast, CC-GVAE performs poorly in target accuracy for both activation and deactivation.

In essence, unlike the strong gains reported for steerable image generation, CB-GVAE and CC-GVAE do not consistently outperform C-GVAE on molecular graphs under single-concept interventions. However, CB-GVAE achieves substantial gains under minimum Hamming distance interventions, particularly for large concept sets. We discuss plausible reasons for these findings, and directions for improvement, in the section below.

\begin{table}  
\centering
\caption{Target and non-target accuracy under concept activation and deactivation.}
\label{tab:concept_steerability}
\resizebox{\textwidth}{!}{%
\begin{tabular}{llcccccccc}
\toprule
\multirow{2}{*}{Concept Set} & \multirow{2}{*}{Model}
& \multicolumn{2}{c}{Activation}
& \multicolumn{2}{c}{Deactivation}
& \multicolumn{2}{c}{Hamm. Activation}
& \multicolumn{2}{c}{Hamm. Deactivation} \\
\cmidrule(lr){3-4}
\cmidrule(lr){5-6}
\cmidrule(lr){7-8}
\cmidrule(lr){9-10}
& & Target$\uparrow$ & Non-Target$\uparrow$ & Target$\uparrow$ & Non-Target$\uparrow$ & Target$\uparrow$ & Non-Target$\uparrow$ & Target$\uparrow$ & Non-Target$\uparrow$ \\
\midrule

\multirow{3}{*}{B-10}
& C-GVAE    & $16.12$ & $ \mathbf{78.24} $ & $30.29 $ & $\mathbf{75.63} $ &  $ 18.45 $ & $ 74.49$ & $30.29$ & $74.54$\\
& CB-GVAE   & $\mathbf{31.98} $ & $54.77$ & $\mathbf{47.47}$ & $51.68 $ & $ \mathbf{35.85} $ & $ 55.60 $ & $ \mathbf{51.77} $ & $52.64$ \\
& CC-GVAE   & $18.84 $ & $66.68 $ & $39.77 $ & $64.07 $ & $ 9.02$ & $ \mathbf{85.37}$ & $ 17.84$ & $ \mathbf{83.65}$ \\
\midrule

\multirow{3}{*}{N-10}
& C-GVAE & $25.12 $ & $\mathbf{75.30} $ & $43.78 $ & $\mathbf{71.45} $ & $ 24.34$ & $ 74.91 $ & $43.33 $ & $72.17  $ \\
& CB-GVAE  & $ \mathbf{32.49}$ & $62.88$ & $ \mathbf{56.74} $ & $57.49$ & $\mathbf{33.82}$ & $61.50 $ & $\mathbf{57.11}$ & $57.08 $ \\
& CC-GVAE   & $20.03$ & $67.69 $ & $45.16 $ & $62.75 $ & $ 8.75 $ & $ \mathbf{85.73} $ & $19.43 $ & $ \mathbf{82.95} $ \\
\midrule

\multirow{3}{*}{B-35}
& C-GVAE   & $15.40$ & $\mathbf{79.20} $ & $27.43$ & $ \mathbf{78.00}$ & $ 24.68	 $ & $66.36 $  & $ 39.14 $ & $66.92$ \\
& CB-GVAE  & $17.83 $ & $71.36$ & $27.26$ & $71.18 $ & $ \mathbf{31.51}  $ & $ 49.34 $ & $ \mathbf{49.23} $ & $ 46.25$ \\
& CC-GVAE  & $\mathbf{19.03} $ & $67.84 $ & $ \mathbf{35.46} $ & $65.20 $ & $ 9.15$ & $ \mathbf{85.25} $ & $ 17.30 $ & $ \mathbf{82.00}$ \\
\midrule

\multirow{3}{*}{N-35}
& C-GVAE    & $\mathbf{15.06} $ & $\mathbf{78.99}$ & $ \mathbf{44.63} $ & $ \mathbf{78.33} $ & $18.85 $ & $ 73.58$ & $48.66 $ & $75.36$ \\ 
& CB-GVAE   & $13.52$ & $72.73$ & $31.98$ & $78.17$ & $ \mathbf{25.11} $ & $ 56.25$ & $ \mathbf{55.70} $ & $ 57.71 $ \\
& CC-GVAE   & $13.30 $ & $70.72$ & $41.78 $ & $69.13 $ & $ 5.21  $ & $ \mathbf{87.32} $ & $ 19.60$ & $ \mathbf{85.91}$ \\

\bottomrule
\end{tabular}%
}
\end{table}


\begin{figure}
    \centering
    \includegraphics[width=0.47\textwidth]{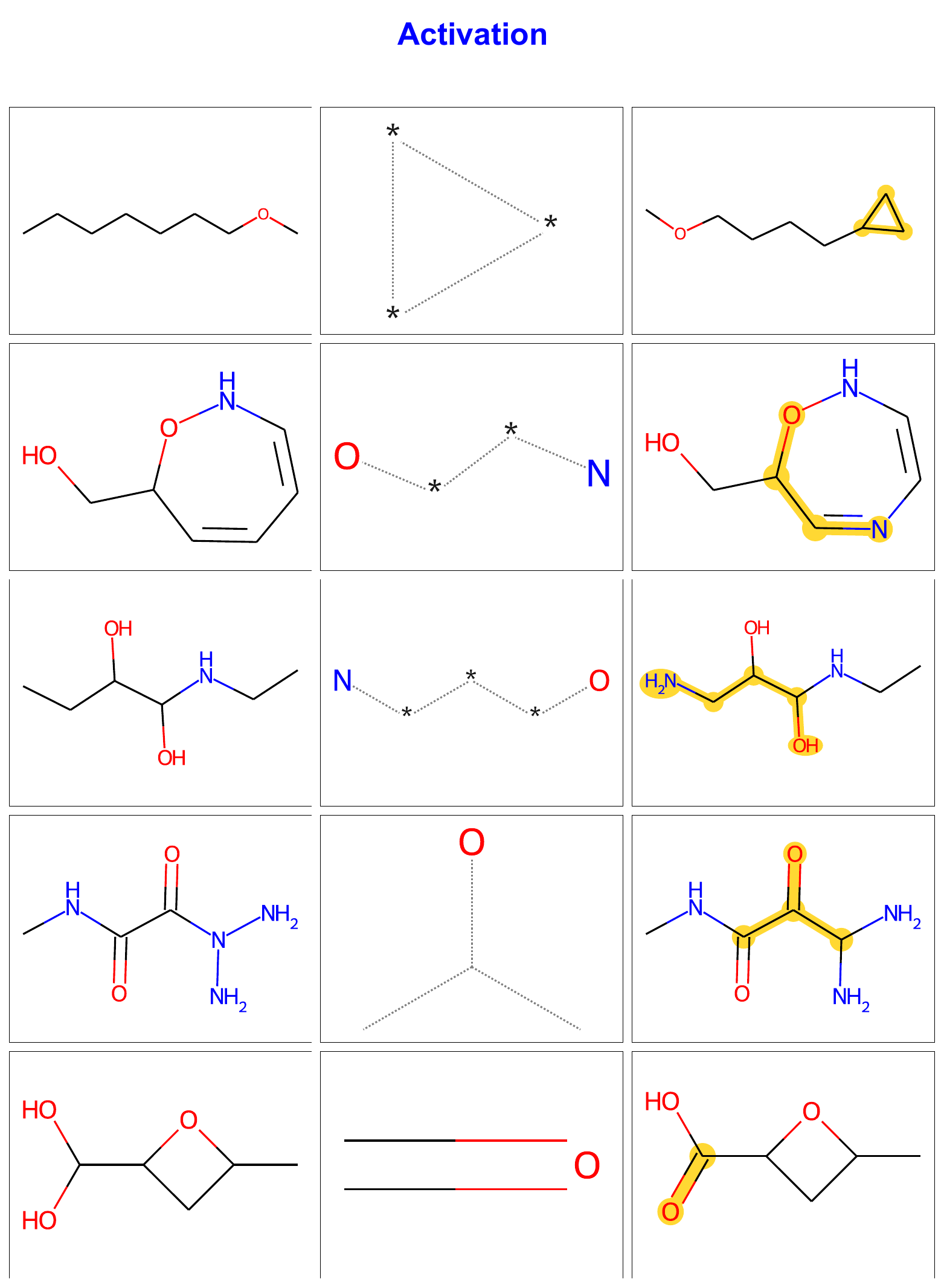}
    \includegraphics[width=0.47\textwidth]{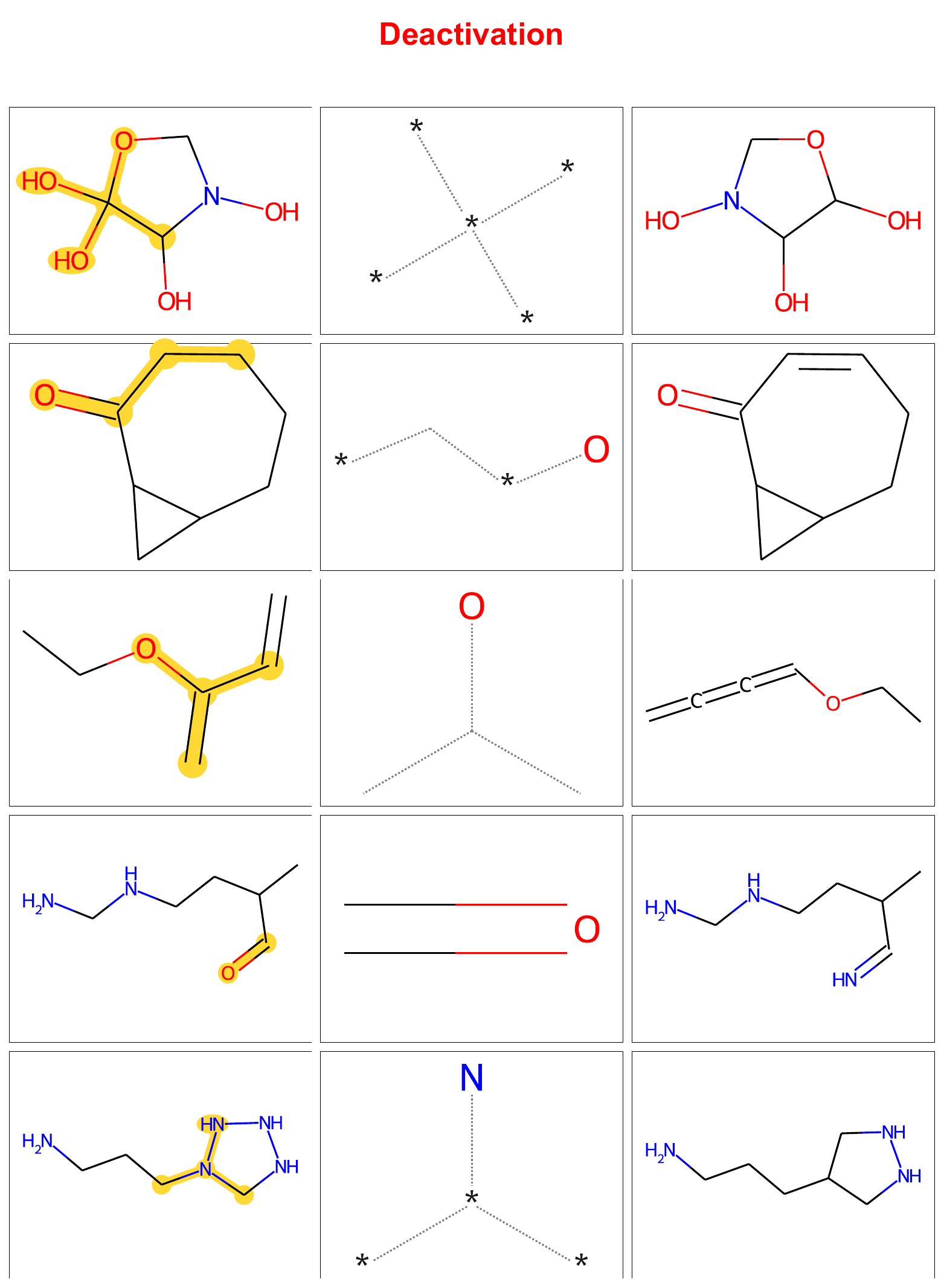}
    \caption{\textbf{Steering with CB-GVAE}. The left and right panels show concept \textit{activation} and \textit{deactivation}, respectively. In each row: the left box shows the generated molecule, the middle box shows the targeted substructure, and the right box shows the molecule after intervention, with the substructure highlighted. Dotted lines denote bonds of unspecified type, where a vertex joined by two or three such bonds is a carbon with implicit hydrogens; asterisks (*) denote wildcard atoms. See Appendix~\ref{appendix:additional_examples} for more examples.}
    \label{fig:activation_deactivation}
\end{figure}

\section{Conclusion}

A plausible explanation for why CB-GVAE and CC-GVAE do not consistently outperform C-GVAE is \emph{concept leakage} within the bottleneck layer, which may stem from our naive design choice of using a simple MLP as the concept encoder. In future work, we plan to replace this with Concept Embedding Models~\cite{espinosa2022concept}, which have been shown to offer more effective concept interventions than simple MLP-based bottlenecks. Additionally, the generative backbone used in this work does not employ graph matching and instead relies on a canonical node ordering over atoms. Incorporating graph matching would render the framework unscalable, since Simonovsky et al.~\cite{simonovsky2018graphvae} note that its matching complexity scales as $O(k^4)$. To address this, we intend to adopt JT-VAE~\cite{jin2018junction}, a scalable generative model, as the backbone in future work. This would allow us to evaluate the generation quality, interpretability, and steerability of concept bottlenecks on larger molecular datasets such as ZINC~\cite{gomez2018automatic}, MOSES~\cite{polykovskiy2020molecular}, and GuacaMol~\cite{brown2019guacamol}.

We present ongoing work adapting Concept Bottleneck Generative Models~\cite{ismail2024concept, martinez2026probabilistic, kulkarni2025interpretable} and Concept Controller~\cite{kulkarni2025interpretable} to molecular graph generation using structural concepts. Although preliminary results show that incorporating a concept bottleneck improves distributional similarity to the reference set (i.e., lower FCD~\cite{polykovskiy2020molecular}) with little to no consistent cost to validity or uniqueness, steerability gains over conditional baselines remain less consistent than reported for images, likely due to concept leakage. We view this work as a foundation for future concept-bottleneck-based one-shot molecular generators that enable interpretability and steerability of generated molecules.

\section{Code Availability}
The source code will be made publicly available upon publication. Researchers interested in accessing the code prior to publication can contact the authors.

\bibliographystyle{plain}
\bibliography{references}

\newpage
\appendix

\section{Reconstruction Loss}
\label{appendix:recon}
Simonovsky et al.~\cite{simonovsky2018graphvae} factorize reconstruction loss $-\log p_\theta(G|z)$ into three separate terms: adjacency, node attributes, and edge attributes, so that
\begin{equation}
-\log p_\theta(G|\mathbf{z}) = -\lambda_A \log p(A|\mathbf{z}) - \lambda_F \log p(F|\mathbf{z}) - \lambda_E \log p(E|\mathbf{z}),
\end{equation}
where $\lambda_A=  \lambda_F = \lambda_E = 1$ are scalar hyperparameters. We fix a node ordering for the molecules using \emph{RDKit}~\cite{landrum2016rdkit} canonical ordering rather than the expensive graph matching formulation proposed by the authors. The ground-truth graph is then simply zero-padded up to the fixed output size of decoder $k$, giving a padded ground-truth adjacency matrix $A'$, a padded node attribute matrix $F'$, and a padded edge attribute tensor $E'$. The reconstruction terms are then given by
\begin{equation}
\begin{split}
\log p(A|\mathbf{z}) = \; & \frac{1}{k}\sum_a A'_{a,a}\log \tilde{A}_{a,a} + (1-A'_{a,a})\log(1-\tilde{A}_{a,a}) \\
& + \frac{1}{k(k-1)}\sum_{a \neq b} A'_{a,b}\log \tilde{A}_{a,b} + (1-A'_{a,b})\log(1-\tilde{A}_{a,b}),
\end{split}
\end{equation}
\begin{equation}
\log p(F|\mathbf{z}) = \frac{1}{n}\sum_i \log F_{i,.}'^\top \tilde{F}_{i,.} \qquad \log p(E|\mathbf{z}) = \frac{1}{\|A\|_1}\sum_{i \neq j} \log E_{i,j,.}'^\top \tilde{E}_{i,j,.}.
\end{equation}

$\|A\|_1$ denotes the sum of all entries in the matrix $A$. It is worth being explicit about the diagonal terms $A'_{a,a}$ versus $\tilde{A}_{a,a}$: $A'_{a,a}$ is a fixed ground-truth label, equal to $1$ if slot $a$ is a real atom in the molecule and $0$ if it falls in the padded region. $\tilde{A}_{a,a}$ is the probability that node $a$ exists, and is implictly predicted from the edges rather than output directly: $\tilde{A}_{a,a} = \max_b \tilde{A}_{a,b}$, the maximum predicted probability over all edges incident to node $a$.

\section{MACCS Concept Bits}
\label{appendix:concept_bit_list}
\begin{table}[h]
\centering
\caption{MACSS concepts bits used for each subset}
\label{tab:subsets}
\small
\begin{tabular}{l p{0.8\textwidth}}
\toprule
\textbf{Subset} & \textbf{Concept Bits} \\
\midrule
B-10 & 96, 105, 121, 132, 151, 152, 153, 156, 158, 159 \\
\addlinespace
\midrule
\addlinespace
N-10 & 11, 22, 96, 104, 109, 127, 150, 158, 160, 163 \\
\addlinespace
\midrule
\addlinespace
B-35 & 11, 22, 57, 66, 83, 90, 91, 95, 96, 97, 104, 105, 109, 112, 116, 121, 127, 128, 129, 132, 139, 143, 150, 151, 152, 153, 154, 155, 156, 157, 158, 159, 160, 161, 163 \\
\addlinespace
\midrule
\addlinespace
N-35 & 11, 17, 19, 22, 38, 41, 43, 53, 69, 71, 74, 75, 76, 86, 89, 92, 93, 94, 96, 99, 101, 102, 104, 106, 109, 114, 123, 127, 136, 138, 141, 147, 158, 160, 163 \\
\bottomrule
\end{tabular}
\end{table}



\section{Conditional Graph Variational Autoencoder (C-GVAE)}
\label{appendix:c-gvae}
Following the conditioning strategy introduced by Simonovsky et al.~\cite{simonovsky2018graphvae}, we condition both the encoder and the decoder on the label vector $y$ associated with each input graph $G$. Specifically, the decoder $p_\theta(G \mid \mathbf{z},\mathbf{y})$ receives the concatenation of $z$ and $y$, while in the encoder $q_\phi(\mathbf{z}\mid G,\mathbf{y})$, $\mathbf{y}$ is concatenated with the features of every node before the graph pooling layer.

\newpage
\section{Effect of Epsilon for CC-GVAE}
\label{appendix:epsilon}
\begin{table}[h]
\centering
\small
\caption{Steerability results for CC-GVAE. Values are reported as mean $\pm$ standard deviation over 4 runs.}
\label{tab:concept_steerability_for_cc}
\begin{tabular}{cccccc}
\toprule
\multirow{2}{*}{Concept Set} & \multirow{2}{*}{Epsilon}
& \multicolumn{2}{c}{Activation}
& \multicolumn{2}{c}{Deactivation} \\
\cmidrule(lr){3-4}
\cmidrule(lr){5-6}
& & Target Acc.$\uparrow$ & Non Target Acc.$\uparrow$ & Target Acc.$\uparrow$ & Non Target Acc.$\uparrow$ \\
\midrule

\multirow{2}{*}{B-10}
& 0.2 & $13.26 \pm 2.80$ & $77.73 \pm	4.01$ & $27.27 \pm 1.43$ & $ 74.69 \pm 4.66$ \\
& 0.5 & $18.84 \pm 3.18$ & $66.68 \pm 5.02$ & $39.77 \pm 0.87$ & $ 64.07 \pm 4.69$ \\
\midrule

\multirow{2}{*}{N-10}
& 0.2   & $13.84 \pm 2.83$ & $77.89 \pm 4.49$ & $30.92 \pm 0.66$ & $74.31 \pm 4.39$      \\
& 0.5   & $20.03 \pm 3.85$ & $67.69 \pm 5.61$ & $45.16 \pm 1.69$ & $62.75 \pm 4.66$ \\
\midrule

\multirow{3}{*}{B-35}
& 0.2    & $13.35 \pm	2.06$& $78.10	\pm 3.56 $  & $24.78 \pm 0.89$ & $75.86	\pm 3.84 $ \\
& 0.5   & $19.03 \pm 1.90$ & $67.84	\pm 4.08$ & $35.46 \pm 0.78$ & $65.20 \pm 4.78$ \\
\midrule

\multirow{3}{*}{N-35}
& 0.2   & $8.46  \pm 1.48$ & $80.27	\pm 3.40$ & $29.49 \pm 3.35$ & $78.38 \pm 3.67$\\
& 0.5   & $13.30 \pm 1.58$ & $70.72	\pm 4.26$ & $41.78 \pm 1.71$ & $69.13 \pm 4.17$\\

\bottomrule
\end{tabular}
\end{table}

\section{Additional Examples}
\label{appendix:additional_examples}
\begin{figure}[h]
    \centering
    \includegraphics[width=0.42\textwidth]{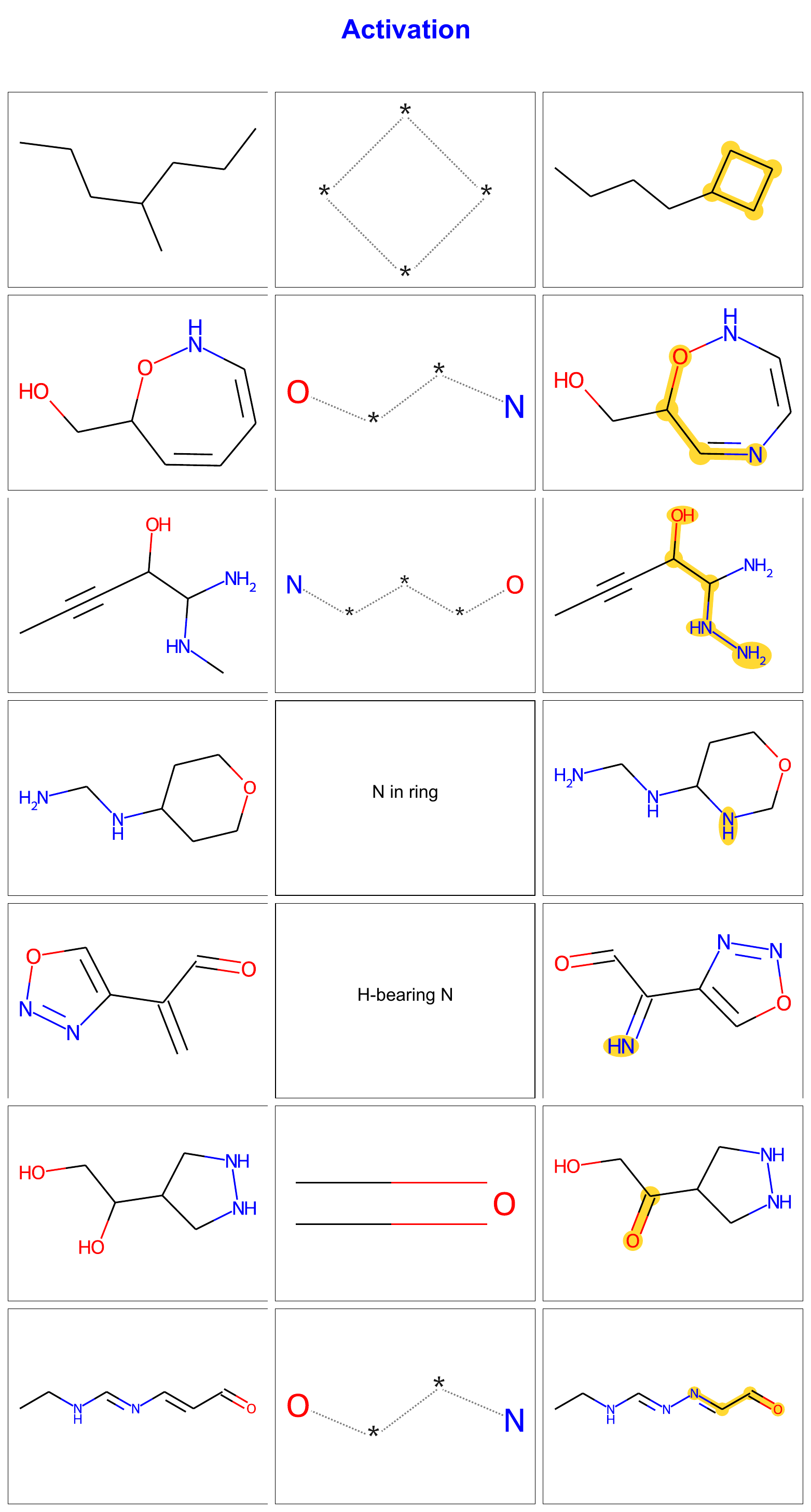}
    \includegraphics[width=0.42\textwidth]{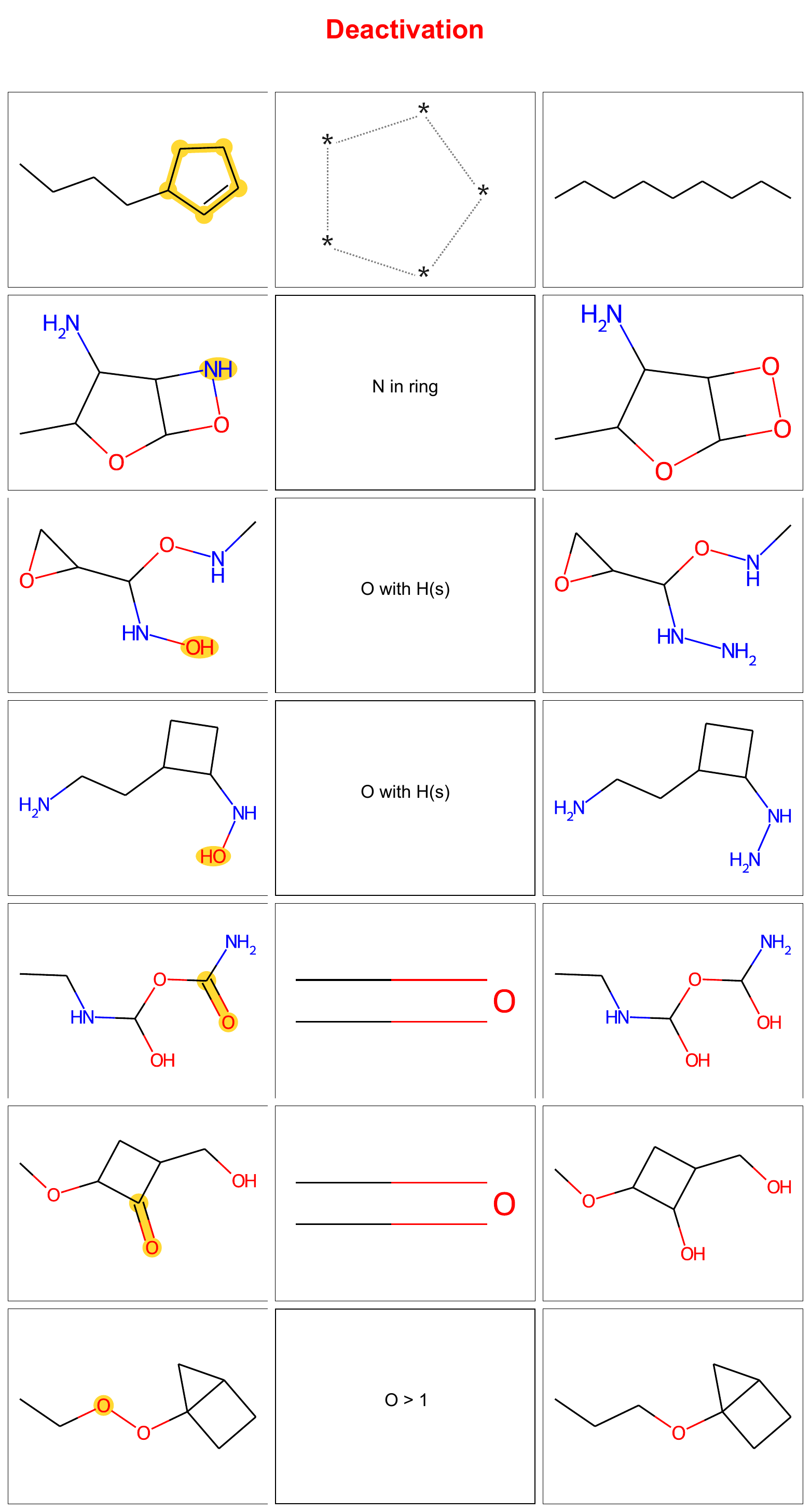}
    \caption{Additional qualitative results using CB-GVAE (B-35)}
    \label{fig:additional_steerability_results}
\end{figure}

\section{Sample Generations}
\label{appendix:random_samples}
\begin{figure}[h]
\centering
\includegraphics[width=0.85\textwidth]{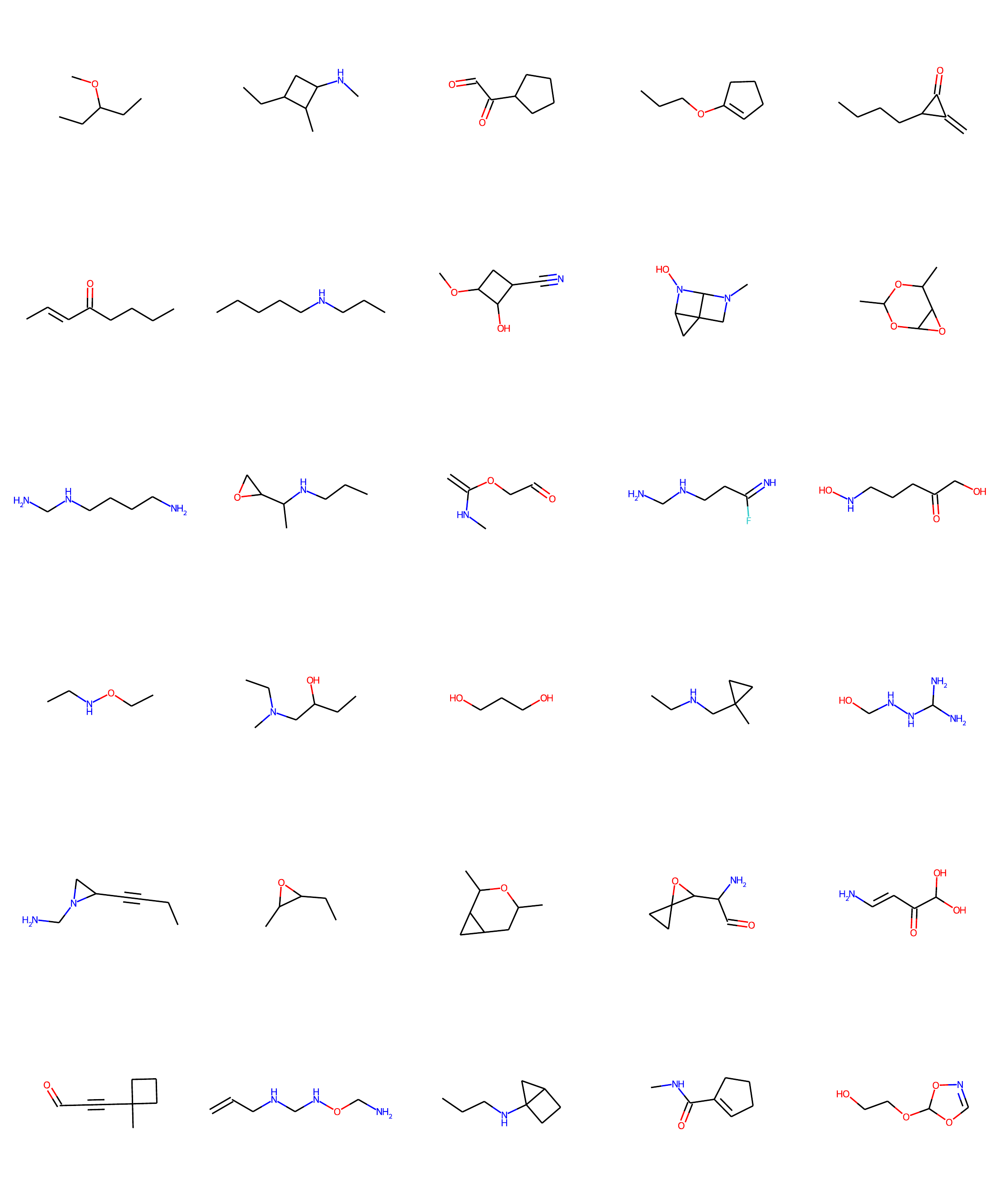}
\caption{Samples from CB-GVAE (B-35)}
\label{fig:samples_mac35}
\end{figure}

\begin{figure}[h]
\centering
\includegraphics[width=0.85\textwidth]{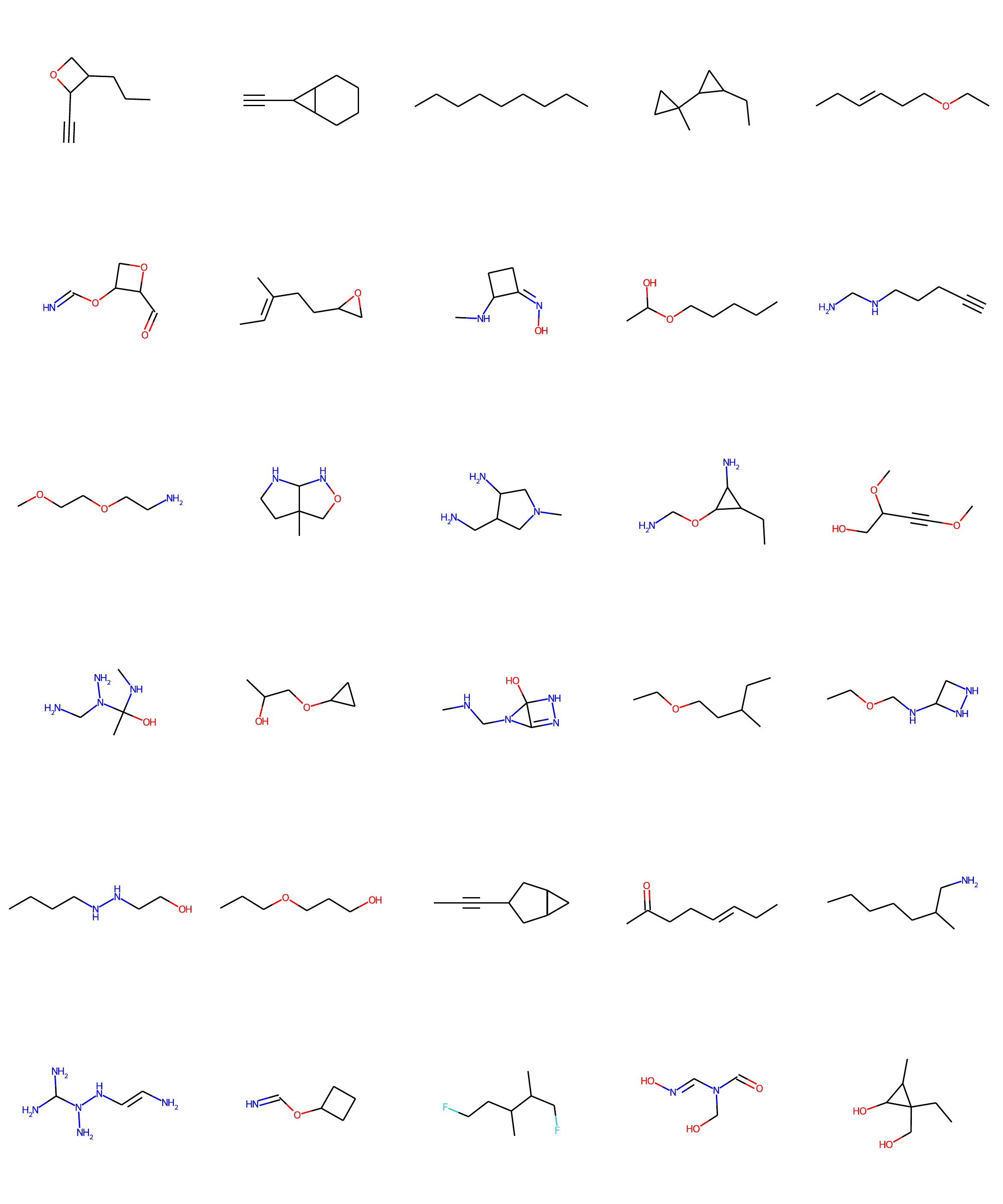}
\caption{Samples from GVAE}
\label{fig:samples_vae}
\end{figure}

\end{document}